\documentclass[11pt]{article}

\usepackage[]{acl}

\usepackage{times}
\usepackage{latexsym}
\usepackage{amssymb}

\usepackage[T1]{fontenc}

\usepackage{subcaption}

\usepackage[utf8]{inputenc}

\usepackage{microtype}

\usepackage{booktabs}
\usepackage{xcolor}
\usepackage{rotating}
\usepackage{multirow}

\usepackage{todonotes}
\usepackage{caption}
\usepackage{subcaption}
\usepackage{xspace}

\usepackage{graphicx}

\newcommand{\sep}{$\langle${\sc sep}$\rangle$\xspace}
\newcommand{\boundary}{\ensuremath\mid\xspace}

\newcommand{\paralleldata}{\ensuremath{\cal P}}
\newcommand{\btdata}{\ensuremath{\cal B}}
\newcommand{\doc}{\ensuremath{_d}}

\newcommand{\sys}[1]{{\sc #1}}
\newcommand{\syssent}{\sys{Sent}(\paralleldata,\btdata)\xspace}
\newcommand{\syssentdoc}{\sys{Sent$\star$}(\paralleldata,\btdata)\xspace}
\newcommand{\sysdocboth}{\sys{Doc}(\paralleldata\doc,\btdata\doc)\xspace}
\newcommand{\sysdocpar}{\sys{Doc}(\paralleldata\doc,\btdata)\xspace}
\newcommand{\sysdocbt}{\sys{Doc}(\paralleldata,\btdata\doc)\xspace}
\newcommand{\ellvp}{ell$_{\mathrm{VP}}$}
\newcommand{\ellinfl}{ell$_{\mathrm{infl}}$}

\newcommand{\deen}{DE$\rightarrow$EN\xspace}
\newcommand{\engdeu}{EN$\rightarrow$DE\xspace}
\newcommand{\enfr}{EN$\rightarrow$FR\xspace}
\newcommand{\enru}{EN$\rightarrow$RU\xspace}

\newcommand{\repo}{\url{https://github.com/marian-nmt/docmt23}}

\title{Why are we still translating sentences?}
\title{Three changes for escaping the sentence-level paradigm in machine translation}
\title{Escaping the sentence-level paradigm in machine translation}

\definecolor{mjdpurple}{RGB}{181, 124, 193}

\author{Matt Post \and Marcin Junczys-Dowmunt \\
  Microsoft \\
  Redmond, Washington \\
  \texttt{\{mattpost,marcinjd\}@microsoft.com}
}

\begin{document}
\maketitle
\begin{abstract}
It is well-known that document context is vital for resolving a range of translation ambiguities, and in fact the document setting is the most natural setting for nearly all translation.
It is therefore unfortunate that machine translation---both research and production---largely remains stuck in a decades-old sentence-level translation paradigm.
It is also an increasingly glaring problem in light of competitive pressure from large language models, which are natively document-based.
Much work in document-context machine translation exists, but for various reasons has been unable to catch hold.
This paper suggests a path out of this rut by addressing three impediments at once: what architectures should we use? where do we get document-level information for training them? and how do we know whether they are any good?
In contrast to work on specialized architectures, we show that the standard Transformer architecture is sufficient, provided it has enough capacity.
Next, we address the training data issue by taking document samples from back-translated data only, where the data is not only more readily available, but is also of higher quality compared to parallel document data, which may contain machine translation output.
Finally, we propose generative variants of existing contrastive metrics that are better able to discriminate among document systems.
Results in four large-data language pairs (\deen, \engdeu, \enfr, and \enru) establish the success of these three pieces together in improving document-level performance.
\end{abstract}

\section{Introduction}

There are two key components to the remarkable advances in the field of Natural Language Processing over the past few years.
The first of these is the architecture, which is the original Transformer \cite{vaswani-etal-2017-attention} with a number of incremental tweaks, but importantly, scaled to larger sizes.
The second is the data: training on more and more of it, and extending the basic unit from the sentence to the document.
Encoder-only models such as BERT \cite{devlin-etal-2019-bert} allowed up to 512 tokens of context, bringing it well over typical length limits employed for machine translation even today.
Decoder-only large language models (LLMs) extended this to thousands of tokens \cite{brown-etal-2020-language}, and make use of documents as their basic unit of training.

\begin{figure}
    \centering
    \includegraphics[width=0.45\textwidth]{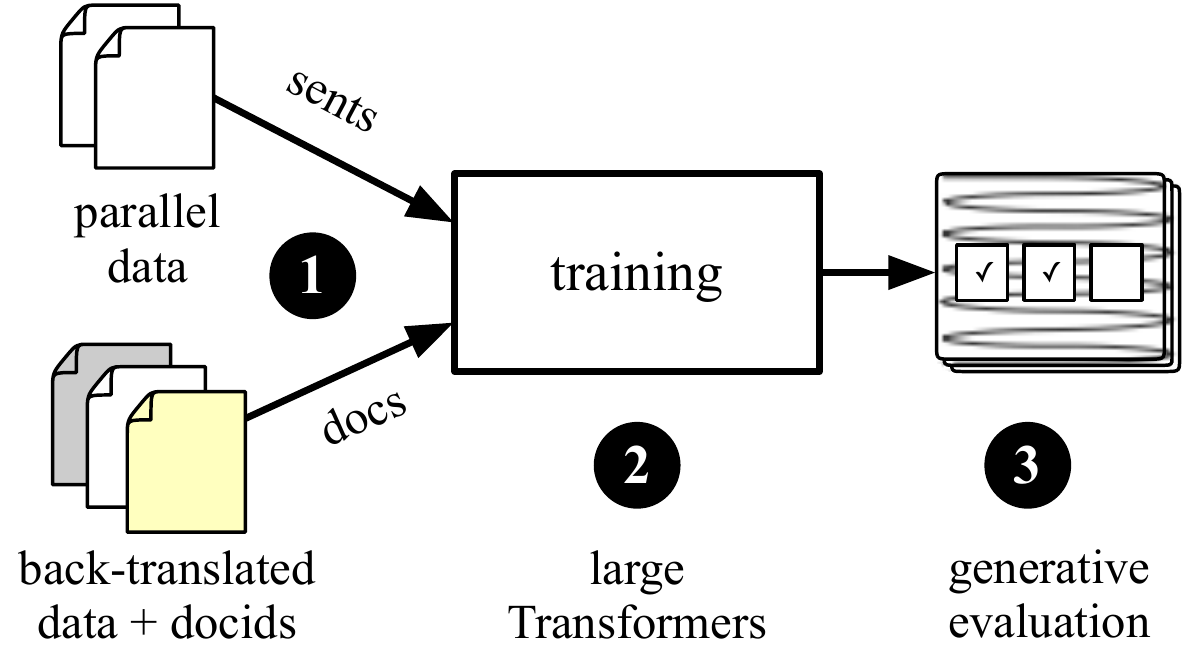}
    \caption{Escaping the rut of sentence-level translation: (1) source documents from trustworthy data only, (2) feed them into large-capacity standard Transformer models, and (3) use test sets that evaluate a model's generative ability.}
    \label{figure:page1}
\end{figure}

Compared to the advances in large language models, machine translation is something of an outlier.
Instead of simple architectures and document-level data, we make little use of document-level data, with work often focused on complex extensions to the Transformer architecture (e.g., \citet{lopes-etal-2020-document,yu-etal-2020-better}).
There are many sensible reasons for this.
The most immediate explanation is that existing training data does not have the document metadata that is needed for document training, placing an impediment at the very start of any effort.
Although all training data originates in documents, the typical extraction pipeline drops this information in the cleaning and deduplication process, and large dataset releases remain sentence-based.
In fact, in an attempt to find every last parallel morsel for training, recent large extraction projects have doubled down on this approach, using sentence-level extraction and alignment techniques that preclude document-level information entirely \cite{banon-etal-2020-paracrawl}. 
Revising these extraction processes presents an immense amount of work.
A second reason is that for most languages of interest, sentences provide a reasonable base unit of meaning, and translating them in parallel helps keep operational costs and latencies low. 
A third reason is the difficulty of evaluation: document-level phenomena are rare enough that gains there are unlikely to be reflected with current test-set-level metrics, leading to a perception, perhaps, of diminishing returns, and to the idea that the effort is not worthwhile when all costs are considered.
You can't improve what you can't measure.

Together, these problems present a steep rut: without good evaluation, we cannot 
know if there are worthwhile improvements, and without good data and models, we cannot achieve them.
This leaves us stuck in a sentence-level paradigm whose age is showing every day, particularly in light of work demonstrating the document-level translation abilities of large language models \cite[LLMs]{zhang-etal-2023-prompting,hendy-etal-2023-how,karpinska-iyyer-2023-large,wang-etal-2023-document-level}.
Despite significant prior work on the topic (\S~\ref{section:related-work}), and general acknowledgment of the need to move on \cite{sennrich-2018-ripe}, document-context translation has not managed to take hold in the field, and may be its most unaddressed problem.

This paper presents findings that we hope can help move machine translation into a document-based paradigm, even without the help of LLMs.
It encompasses three main findings.
We
\begin{itemize}
    \item show that document-based translation works well with the standard Transformer architecture, provided it has sufficient capacity;
    \item demonstrate that training with document samples sourced from backtranslated data \emph{alone} works well, and moreover, that parallel data as a whole is even harmful; and
    \item show that contrastive test sets are not able to discriminate document-based systems, and introduce generative variants (\S~\ref{section:evaluation}) that do.
\end{itemize}

\noindent Critically, it seems to be all three changes at once that enable quality document translation models.

\section{The need for document-level translation}

\begin{table*}[th]
    \centering
    \begin{tabular}{p{.1cm}p{.3\textwidth}|p{.65\textwidth}}
    \toprule
    & \textbf{phenomenon}
    & \textbf{example}
    \\    

    \midrule
    i
    & \emph{Anaphora}, words whose meaning or selection depends on another word in context (especially, pronouns)
    & \emph{Is there a rise of \textbf{anti-Semitism} in America as well in the last ten, twenty years? No. If anything, \textbf{it} has declined. $\rightarrow$ Gibt es auch in Amerika eine Zunahme des \textbf{Antisemitismus} in den vergangenen zehn, 20 Jahren? Nein. \underline{Er} ist eher zurückgegangen.}
    \\
    \midrule
    ii
    & \emph{Deixis}, words whose interpretation requires (usually) extra-linguistic content such as the speaker identity, place, or time
    & \emph{Thank you, \textbf{Your Majesty}. It's a pleasure to meet \textbf{you}. $\rightarrow$ Merci, \textbf{votre Majesté}. C'est un plaisir de {\underline{\color{red}te}} recontrer.} The context precludes the use of the informal \emph{tu}.
    \\
    \midrule
    iii
    & \emph{Ellipsis}, the omission of text not needed in the source
    & \emph{Pensaba que habías dejado de correr. $\rightarrow$ I/he/she thought you stopped running.} Context is required to determine the correct subject.
    \\

    \midrule
    iv
    & \emph{Discourse connectives}, which cue readers to higher level document structure
    & \emph{La liberté ne se décrète pas. \textbf{Or}, la construction communautaire ne semble pas consciente de cette réalité.} (``Liberty cannot be decreed. \textbf{But} those who are building Europe seem unaware of that fact.'')
    As explained in \citet{zufferey-2016-discourse}, without context, the translation of the French connective \emph{or} changes whether sentence concedes a point to the main argument (as here, ``But'') or continues it (``In fact'').
    \\

    \midrule
    v
    & \emph{Grammatical and lexical cohesion}, a surface property related to word selection and sentence structure
    & \emph{What's \textbf{crazy} about me? Is this \textbf{crazy}? $\rightarrow$ Qu'est-ce qu'il y a de dingue chez moi~? Est-ce que ça c'est {\color{green}\textbf{dingue}} [{\color{red}\textbf{fou}}]~?}
    The emphatic context requires that \emph{crazy} be translated consistently \cite{bawden-etal-2018-evaluating}
    \\
    \end{tabular}
    \caption{Discourse phenomena taxonomy and definitions adapted from \citet[Table 1, p.9]{maruf-etal-2019-survey}.
    Here, we arrange them along a rough spectrum moving from adequacy (i) to fluency (v), and add examples.}
    \label{table:phenomena}
\end{table*}

Production and consumption of written text largely occurs at the paragraph or document level.
While many sentences can be understood in isolation, there are many whose meaning cannot be perfectly understood unless it is evaluated in this context.
This is important for translation, because in addition to the contextual nature of understanding, different languages map from meaning to surface form in different ways, and there are consequently a range of issues introduced depending on the particular language pair and setting.
In Table~\ref{table:phenomena}, we summarize a number of these phenomena, adapting Table~1 of \citet{maruf-etal-2019-survey}, reorganizing a bit and adding examples.\footnote{We drop \emph{coherence} as a separate entry, in the belief that it is largely a monolingual property of the source text \cite[Ch.21]{jurafsky-martin-2008-speech}, and in MT is consequently difficult to distinguish from cohesion.}
Anaphora (i), deixis (ii), ellipsis (iii), and the selection of appropriate connectives (iv) are mostly related to adequacy, since they require the translation system to resolve semantic ambiguities that are introduced at the sentence level.
Cohesion (v) pertains more to fluency issues, since it involves choices mostly dealing with the quality of the target-language text.  %
This boundary between adequacy and fluency is not always clear, however.
The basic point is that so long as sentences are translated in isolation, most of these issues cannot be resolved without extra-sentential information.
The sentence-level paradigm therefore places a ``parallel ceiling'' on translation quality, one that fundamentally limits the abilities of systems operating within it.
The exact height of this ceiling is difficult to state in the general case.
Many document-level phenomena are relatively rare, and the success of commercial MT suggests that the errors inherent in the paradigm may be long-tail obstacles not worth addressing with new complications to architectures or training pipelines.
It is not a focus of this paper, but we address the question of the ``discourse-density'' of texts somewhat in Section~\ref{section:dense}.

Document-context MT may also be useful in other settings.
Headlines, for example, typically have their own style, and in our observations, docMT systems seem better-equipped to notice and preserve it, even without explicit tags about document structure.
Languages such as Thai do not mark end-of-sentence, making pre-processing difficult, which could be solved by working at the paragraph level.
We also note that many tasks that currently require specialized approaches might naturally be subsumed by document-context translation.
One such example is formality \citep{sennrich-etal-2016-controlling}, where enough source-side context might help establish the appropriate target-side register, and where generating targets at the document level would help ensure consistency across sentences.
Another is gender disambiguation.

A fundamental problem for document MT is the difficulty of evaluation.
In the presence of the potentially sparse-nature of document phenomena, and in the absence of proven, general metrics for identifying them, we turn instead to a handful of annotated test sets that previous research has produced.
This limits evaluation to the set of phenomena we have annotations for, which we now discuss.

\section{The challenge of evaluation}
\label{section:evaluation}

An important hurdle in the path to document-level translation is the difficulty of evaluation.
What we really want to know is whether document-context systems improve performance on the phenomena that we expect them to model.
The frequency of these phenomena in standard corpora is not known, however, and we expect them to be relatively rare, such that improvements might not be observable from corpus-level automatic metrics.
Attempts to automatically identify sentences requiring context have shown the task to be difficult \cite{bawden-etal-2018-detecting}.
This makes it challenging to measure improvements on arbitrary corpora.

Fortunately, there exist a range of test sets that have been developed to capture document-level phenomena.
By and large, these test sets are \emph{contrastive} ones, that use model score to determine how well a system can discriminate between good and bad examples.
We begin by cataloguing those that we make use of in this paper (\S~\ref{section:contrastive}).
We then describe a generative extension that makes better use of these contrastive test sets in (\S~\ref{section:generative}).

\subsection{Contrastive test sets}
\label{section:contrastive}

The dominant paradigm for evaluation of long-tail document phenomena has been so-called \emph{contrastive evaluation}, in which a system is tested on its ability to discriminate between correct and incorrect translation pairs.
The correct examples are usually taken from found text; the incorrect ones are created by creating an error of some sort.
Systems are evaluated on the percentage of time they correctly score the positive example above its incorrect variant, by way of model score.
Table~\ref{table:metric-examples} contains examples of each test set.

\paragraph{ContraWSD (DE-EN)}
\citet{rios-gonzales-etal-2017-improving} constructed a dataset consists of 7,359 examples of ambiguous German words in the context of translated books, news articles, Global Voices, and UN assembly transcripts.
To create a contrastive example, a new version of the target side is created by replacing the correct translation sense of that source word with an incorrect sense.
Although not designed as a document-level test bed, word-sense disambiguation (WSD) is often viewed as a task that could benefit from document context.

\paragraph{ContraPro (EN-DE)}
In this dataset, \citet{muller-etal-2018-large} focus on the German pronouns \emph{er}, \emph{es}, or \emph{sie}.
They pair sentences containing naturally-found instances of pronouns drawn from OpenSubtitles \cite{lison-tiedemann-2016-opensubtitles2016} with two variants that are identical except that the correct pronoun has been replaced with each of the two incorrect ones.
The dataset is balanced such that there are 4k correct examples for each of the three German pronouns; each is paired with two incorrect examples, yielding a test set size of 24k.

\paragraph{ContraPro (EN-FR)}
In the course of evaluating a number of metrics for document MT, \citet{lopes-etal-2020-document} introduced an extension of the EN-DE ContraPro for EN-FR.
Its examples are also drawn from OpenSubtitles, but since French has only two genders, there is only one contrastive pair per found instance (contrastive pronouns retain the grammatical number of their counterpart).
The dataset contains 14k examples, evenly split among \emph{il}, \emph{elle}, \emph{ils}, and \emph{elles}.

\paragraph{DiscEvalMT (EN-FR)}
\citet{bawden-etal-2018-evaluating} focuses on both anaphora and lexical disambiguation and consistency, using a set of relatively short, hand-crafted two-sentence examples where the correct word or phrase in translation requires information from earlier context.
A key distinguishing feature of this dataset is that the correct choice is often determined by a \emph{target-side} selection.
For lexical choice (200 examples), a source word is used twice, and the test is whether the system prefers a translation that uses the same translation both times (positive example) or alternates (negative example).
For anaphora (200 examples), a source word is translated into each of two target words with opposite genders, and the system must prefer a pronoun use that is consistent with that choice.

\paragraph{GTWiC (EN-RU) \citep{voita-etal-2019-good}}
\emph{Good Translation, Wrong in Context} (hereafter, GTWiC) focuses on a wider range of document-level phenoma, including lexical cohesion (2k instances) and anaphora (3k) as in earlier test sets, but also of the handling of verb selection (500 instances) and morphology (500) in the presence of source-side ellipsis.

\begin{table}[t]
    \begin{subtable}{0.45\textwidth}
        \centering
        \begin{tabular}[t]{p{\textwidth}}
            \toprule
            Eine Berührung mit dem bespornten \textbf{Absatz} machte, daß sein Pferd sich bäumte und dann davon sprengte\dots$\boundary$ 
            A touch of a spurred \{{\color{green}heel},{\color{red}paragraph}\} made his horse first start and rear, and then bound away\dots
            \\
            \bottomrule
        \end{tabular}
        \caption{ContraWSD example.\label{table:example:contrawsd}}
        \vspace{5mm}
    \end{subtable}
    \begin{subtable}{0.45\textwidth}
        \begin{tabular}[t]{p{\textwidth}}
            \toprule
            The prototype has passed every test, sir. It's working.\
            \boundary
            Der Prototyp hat jeden Test erfolgreich durchlaufen, Sir. \{{\color{green}Er},{\color{red}Es},{\color{red}Sie}\} funktioniert.
            \\
            \bottomrule
        \end{tabular}
        \caption{ContraPro example.\label{table:example:contrapro}}
        \vspace{5mm}
    \end{subtable}
    \begin{subtable}{0.45\textwidth}
        \begin{tabular}{p{\textwidth}}
            \toprule
            So you see how bad the implications are. Yes, they are really quite devastating.\
            \boundary
            \emph{Donc tu vois à quel point les \{{\color{blue}implications},{\color{purple}effets}\} sont mauvaises.} Oui, \{{\color{blue}elles},{\color{purple}ils}\} sont vraiment \{{\color{blue}dévastatrices},{\color{purple}dévastateurs}\}.
            \\
            \bottomrule
        \end{tabular}
        \caption{DiscEvalMT example.
        Both translations are correct, but trigger different agreement schemes.}\label{table:example:discevalmt}
        \vspace{5mm}
    \end{subtable}
    \begin{subtable}{0.45\textwidth}
      \begin{tabular}{p{\textwidth}}
      \toprule
        \includegraphics[width=\textwidth]{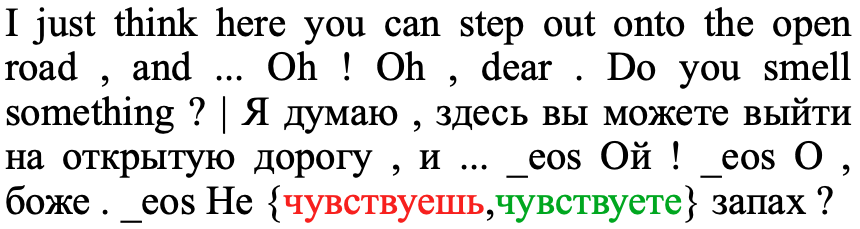} \\
      \bottomrule
      \end{tabular}
      \caption{GTWiC example.
      The first Russian sentence uses the formal register.}    \label{table:example:gtwic}
    \end{subtable}
    \caption{Examples of contrastive test sets.
    The \boundary marks the source/target boundary; \emph{target-side italics} denotes situations where the correct answer depends on a target-side decision (in which case, for generative variants, it is fed as a prefix to the decoder).\label{table:metric-examples}}
\end{table}

\subsection{Testing generative ability}
\label{section:generative}

The challenge sets above test whether a model can discriminate between good and bad examples.
As we will show, many document models perform extremely well on these tasks (Table~\ref{table:source}), but then prove unable to actually generate the correct pronoun at translation time.
The contrastive nature of these test sets was at odds with the actual task: what is needed are metrics that directly evaluate a model's \emph{generative capacity}, rather than its \emph{discriminative ability}.

Fortunately, the datasets described above can all be transformed into generative test sets, with varying degrees of difficulty, depending on how they were assembled.
The core idea is simple: instead of scoring contrastive examples, we (i) translate the source side, (ii) (optionally, where relevant) force-decode the target-side prefix, and then (iii) determine whether the generated content contains the word or phrase in question.
For some datasets (both ContraPros), the correct word is provided as an annotation, which makes this task trivial.
For the others, we infer the correct and incorrect answers by running a phrase-level diff between the positive and negative example in the contrastive pair, which yields the bad and good phrase for each sentence.\footnote{For the English--Russian GTWiC corpus, we first apply the Moses detokenizer, since the dataset was provided tokenized.}

We then produce a generative test score for these test sets as a simple accuracy measure that records success when (a) the positive word or phrase is present in the output and (b) none of the negative words or phrases is present.
Testing takes place at the token level using case-insensitive matching.
Note that this does not work perfectly.
We expect there are ways to get the answer right that do not require direct selection of the word, but trust that large gains reflect real improvements.\footnote{An alternative generative approach is to force-decode the target context prior to the word under question.
We explore this in Section~\ref{section:force-decode}.}

Code for generating and scoring our generative variants can be found at \repo.

\section{Method}
\label{section:method}

\paragraph{Training}

\label{section:constructing-docs}

Our approach to document translation is the simple concatenative document-to-document approach \citep{tiedemann-scherrer-2017-neural,junczys-dowmunt-2019-microsoft}: at training time, we produce training samples by concatenating paired source and target sentences that are within the same document.
Instead of using a fixed pseudo-document length, we concatenate until a maximum token length, or end-of-document, is reached.
Samples are formed from adjacent, non-overlapping sequences of sentences in the training data. 
This is in contrast to the ``multi-resolution'' approach \cite{sun-etal-2022-rethinking}, which creates training samples from many possible sub-sequences of each document, each time the document is observed.
Concatenated sentences are joined with a special \sep token, which ensures that the concatenation is invertible, and also facilitates sentence alignment at inference time.

\paragraph{Inference}
For inference, we employ an overlapping approach.
Each sentence in the input is assembled with left context up to a maximum token length $L$.
The translation system then translates the complete context and the current sentence, which we refer to as the \emph{payload}.
In the output, the \sep token is use to identify the payload's translation.
This is repeated for all sentences in a test set, allowing standard sentence-level metrics to be applied to the results.

\section{Experimental setup}
\label{section:setup}

\subsection{Training Data}

We work with four language pairs: English$\leftrightarrow$German, English$\rightarrow$French, and English$\rightarrow$Russian.
Our data comprises the following sources (Table~\ref{table:data}):
\begin{itemize}
\item Monolingual and parallel data crawled from the web, all containing document metadata.
\item The CCMatrix parallel data release \citep{schwenk-etal-2021-ccmatrix}, which has no document information. 
\end{itemize}

We backtranslate the monolingual data using internal 12-layer sentence-level Transformer systems trained for 20 virtual epochs (\S~\ref{section:models}) on just the parallel data.
Our document-level models then vary based on whether document-level samples are sourced from the backtranslated data, the parallel data, or both.

Ideally, we would conduct experiments on entirely public data.
However, the full breadth of our experiments is not possible in that setting, since document metadata for publicly-released parallel data is rare.
We include a corroborative experiment on open data in Section~\ref{section:wmt}.

\begin{table*}[t]
    \centering
    \scalebox{0.95}{
    \begin{tabular}{l|rrr|rrr|rrr|rrr}
       \toprule
       & \multicolumn{3}{c}{\textbf{German--English}}
       & \multicolumn{3}{c}{\textbf{English--French}}       
       & \multicolumn{3}{c}{\textbf{English--German}}
       & \multicolumn{3}{c}{\textbf{English--Russian}}
       \\
       \midrule
       source 
        & lines & docs & mean
        & lines & docs & mean
        & lines & docs & mean
        & lines & docs & mean
        \\
       \midrule
       mono    
         & 252.5  & 5.5  & 45.3
         & 166.4  & 5.5  & 29.7
         & 205.4  & 7.0  & 29.1
         & 202.7  & 6.5  & 31.1
       \\
       parallel
       \\
       $\rightarrow$ crawled
         & 116.7  & 4.7  & 16.6
         & 123.1  & 3.7  & 33.0
         & 116.7  & 4.7  & 16.6
         &  72.4  & 4.7  & 13.2
       \\
       $\rightarrow$ ccmatrix
         & 45.4   & 0    & - 
         & 65.1   & 0    & -
         & 45.4   & 0    & - 
         &  2.4   & 0    & -
       \\
       \bottomrule
    \end{tabular}}
    \caption{Statistics of the training data used in our experiments (lines and docs in millions). 
    The \emph{mean} column is the mean document length in sentences of documents with $\ge2$ sentences.}
    \label{table:data}
\end{table*}

\subsection{Creating samples}
\label{section:augmentibatch}

A common approach in training machine translation models is to iterate over the entire data in what is called a \emph{data epoch}.

The toolkit we used, Marian \cite{junczys-dowmunt-etal-2018-marian}, splits the data into manageably-sized shards (of 100,000 sentences each). 
Instead of data epochs, replacing it instead with the idea of a \emph{virtual epoch}, defined as observing one billion (1e9) target tokens. Within a shard, the original layout of the data is preserved which allows assembly of consecutively appearing sentences into documents. We do observe document boundaries and do not join sentence across them. If a document is located at a shard boundary and split across two shards we treat it as two separated documents which results in some negligible loss.

Since our approach to document context relies on assembling neighboring lines into (pseudo)documents resulting again in one line of text, standard batching methods can be used that would work with any other line-based data source.

Training then proceeds as follows.
We define two pools of data: the parallel data, \paralleldata, and our back-translated data, \btdata, each of which is separately sharded.
Associated with each pool is a flag, $d$, denoting whether documents can be drawn from it, e.g., \paralleldata$_d$ and \btdata$_d$ (the default is to draw only sentences).
We draw training samples from these pools at a 1:1 ratio according to the following process:
\begin{enumerate}
    \item Randomly choose a pool (\paralleldata or \btdata); %
    \item If there is no active shard for the pool, randomly choose a shard;
    \item If the document flag, $d$, is enabled on the pool, choose a sample length $l$ uniform at random from $1{\dots}L$, for some maximum length $L$, and merge sentences on source and target side until the source side reaches this length (in tokens) or the end of the document;
    \item Otherwise, return the next sentence pair.
\end{enumerate}
Note that the position within each shard is retained across queries, so that the shard itself will be completely consumed, once selected.
Marian uses read-ahead to build a large set of such samples, from which they are sorted into batches., which are then processed in random order. %

\subsection{Models}
\label{section:models}

All of our models are Transformers \cite{vaswani-etal-2017-attention} trained with Marian \cite{junczys-dowmunt-etal-2018-marian-cost}.
Our experiments with different model capacities (\S~\ref{section:capacity}) led us to a standard Transformer model with a 12-layer encoder, a 6-layer decoder, an embedding dimension of 1,024, and a feed-forward network size of 16,384.
We train for 40 virtual epochs.\footnote{We define a virtual epoch as updates from one billion target-side tokens.}
We use a batch size of 500k target-side tokens.
Our maximum document sample length is $L=250$ tokens.

We compare the following models, using the syntax \sys{Name}(pool$_1$, pool$_2$) to denote the pools (\S~\ref{section:augmentibatch}) of data each draws from:
\begin{itemize}
    \item \syssent. A sentence-level baseline.
    \item \syssentdoc. A deficient setting that takes the sentence-level baseline and tests it with document-context inputs.\footnote{In this setting alone, no \sep token is used when combining sentences, since the sentence model has not seen them.} %
    \item \sysdocboth. A document-level system, with documents drawn from both parallel and back-translated data.
    \item \sysdocpar. A document-level system, with documents drawn only from parallel data.
    \item \sysdocbt. A document-level system, with documents drawn only from backtranslated data.
\end{itemize}

\begin{table}[t]
    \centering
    \begin{tabular}{l|rr}
      \toprule
      test set & \# lines & \# docs \\
      \midrule
      wmt22:de-en & 1,984 & 271 \\
      wmt22:en-de & 2,037 & 181 \\
      wmt15:en-fr & 1,500 &  76 \\
      wmt22:en-ru & 2,037 & 181 \\
      \bottomrule
    \end{tabular}
    \caption{Test set statistics from WMT15 \cite{bojar-etal-2015-findings} and WMT22 \cite{kocmi-etal-2022-findings}.}
    \label{table:wmt-test-sets}
\end{table}

\subsection{Evaluation}

In addition to the contrastive and generative document-level test suites described in Section~\ref{section:evaluation}, we compute the following COMET\footnote{COMET version 1.1.3 with model \texttt{wmt20-comet-da}; we multiply COMET scores by 100 for readability.} \cite{rei-etal-2020-comet} and sacrebleu\footnote{Signature: \texttt{nrefs:1|case:mixed|eff:no|
tok:13a|smooth:exp|version:2.3.1}} \cite{papineni-etal-2002-bleu,post-2018-call} scores on a WMT test set (Table~\ref{table:wmt-test-sets}).

\section{Experiments}
\label{section:results}

We begin by establishing baseline scores on standard corpus-level metrics when translating with each model at the sentence level (Table~\ref{table:bleu}).
In addition to a commercial baseline (Microsoft, accessed via API), we present results when translating at both the sentence and document levels.
The table establishes a number of findings:
\begin{itemize}
    \item All models are good, state-of-the-art models when evaluated at the sentence level;
    \item We observe a fairly consistent gain of roughly a COMET\footnote{A reminder from \S~\ref{section:evaluation}: COMET 1.1.3 with model \texttt{wmt20-wmt-da}.} point when moving from the baseline sentence-level translation with \syssent (first row top sent-level section) to \sysdocbt;
    \item We do not see consistent improvements in COMET scores when adding context to the \sysdocbt system (or other doc systems)
\end{itemize}

It seems that training with extended context improves the systems' ability to translate at the sentence level, but on these test sets, there are not widespread gains from employing context.

\begin{table*}[t]
    \centering
    \begin{tabular}{ll|rr|rr|rr|rr}
       \toprule
       \multicolumn{2}{c}{} 
       & \multicolumn{2}{c}{\textbf{DE$\rightarrow$EN}}
       & \multicolumn{2}{c}{\textbf{EN$\rightarrow$DE}}
       & \multicolumn{2}{c}{\textbf{EN$\rightarrow$FR}}
       & \multicolumn{2}{c}{\textbf{EN$\rightarrow$RU}}
       \\
       \multicolumn{2}{c}{model}
       & BLEU & COMET
       & BLEU & COMET
       & BLEU & COMET
       & BLEU & COMET
       \\
      \midrule
      \multicolumn{2}{c|}{Microsoft}
       & 33.5 & 55.2 %
       & 37.3 & 62.0 %
        &  40.8    & 67.6  %
       & 33.1 & 67.3 %
      \\
      \midrule
     \multirow{4}{*}{\rotatebox{90}{sent-level}}
      & \syssent
        & 32.8	& 54.5	& 37.2	& 61.6	& 45.6	& 69.0	& 34.0	& 70.0 \\
      & \sysdocboth 
        & 33.0	& 54.9	& 37.5	& 62.0	& 45.1	& 70.0	& 34.1	& 70.4 \\
      & \sysdocpar
	& 32.3	& 54.4	& 37.0	& 61.3	& 45.4	& 69.2	& 33.5	& 70.0 \\
      & \sysdocbt
	& 33.3	& \textbf{55.8}	& 37.2	& \textbf{62.2}	& 44.5	& 69.8	& 34.3	& \textbf{70.2} \\
      \midrule
      \multirow{3}{*}{\rotatebox{90}{doc-level}}
      & \sysdocboth 
	& 33.2	& 54.5	& 37.9	& 62.1	& 42.5	& 69.1	& 34.3	& 69.2 \\
      & \sysdocpar
	& 33.1	& 54.1	& 37.5	& 62.1	& 43.6	& 67.6	& 33.6	& 68.5 \\
      & \sysdocbt
	& 31.8	& 55.7	& 37.0	& 62.1 & 43.1	& \textbf{70.1}	& 34.1	& \textbf{70.6} \\
      \bottomrule
    \end{tabular}
    \caption{Metric scores on WMT22/WMT15 test sets (Table~\ref{table:wmt-test-sets}) when translating as sentences (top block) and with document context (bottom block).
    Numbers within a column are comparable.
    A main comparison is \sys{Sent}(\paralleldata,\btdata)---the sentence level baseline translating sentences---with \sys{Doc}(\paralleldata,\btdata\doc)---the best doc system, translating as documents.
    }
    \label{table:bleu}
\end{table*}

\begin{table*}[t]
    \centering
    \begin{tabular}{l|rr|rr|rr|rr|rr}
       \toprule
       & \multicolumn{2}{c}{\textbf{DE$\rightarrow$EN}}
       & \multicolumn{2}{c}{\textbf{EN$\rightarrow$DE}}
       & \multicolumn{2}{c}{\textbf{EN$\rightarrow$FR}}
       & \multicolumn{4}{c}{\textbf{EN$\rightarrow$RU}}
     \\
      model 
       & \small C/WSD & \small G/WSD
       & \small C/Pro & \small G/Pro
       & \small C/Pro & \small G/Pro
       & \small \ellinfl & \small G/\ellinfl & \small \ellvp & \small G/\ellvp
       \\
      \midrule
      Literature
       & 71.3  & -  %
       & 70.8  & -  %
       & 83.2  & -  %
       & 76.2  & - &  80.0 & -  %
      \\
      \midrule
      \syssent
      & 97.0  & 76.6 %
      & 50.0  & 33.2 %
      & 71.6  & 22.5 %
      & 51.8  & 24.8  &  19.8   &  4.6 %
      \\
      \midrule
      \syssentdoc
       & 97.0 & 70.5  %
       & 69.0  & 46.3  %
       & 93.1  & 62.3  %
       & 77.0  & 32.8 & 55.0  & 19.2  %
      \\
          \sysdocboth
       & \textbf{97.9} & \textbf{78.1} %
       & 76.5 & 47.8 %
       & \textbf{95.1}	& 62.5
       & 84.2 & 35.8 & \textbf{68.0} & 26.0
        \\
      \sysdocpar
        & 97.8 & 77.8  %
        & 71.6 & 41.9 %
        & 94.3	& 60.4
        & 76.2 & 31.8 & 66.2 & 26.4
        \\
      \sysdocbt
        & 97.4 & 76.7 %
        & \textbf{77.9} & \textbf{70.5} %
        & 94.8	& \textbf{77.3}
        & \textbf{84.6} & \textbf{39.6} & 66.0 & \textbf{28.4}
        \\
      \bottomrule
    \end{tabular}
    \caption{Document contrastive test suites and their generative variants.
    Contrastive scores (C/*) are over the entire dataset in order to compare with the literature, while generative scores are over extra-sentential items only.
    Literature scores are taken from \citet[\deen]{rios-gonzales-etal-2017-improving}, \citet[\enfr,\engdeu]{lopes-etal-2020-document}, and \citet{voita-etal-2019-good}.
    Feeding documents to \syssentdoc (which it wasn't trained on) increases contrastive scores over the sentence baseline and generally brings generative scores within line of doc systems trained with parallel data. 
    }
    \label{table:source}
\end{table*}

\begin{table*}[t]
    \centering
    \begin{tabular}{l|rr|rr||rr|rr}
       \toprule
       & \multicolumn{4}{c}{English--French}
       & \multicolumn{4}{c}{English--Russian}
    \\
       & \multicolumn{2}{c}{anaphora}
       & \multicolumn{2}{c}{lexical}
       & \multicolumn{2}{c}{deixis}
       & \multicolumn{2}{c}{lexical}

      \\
      model 
       & Con. & Gen. & Con. & Gen.
       & Con. & Gen. & Con. & Gen.       
      \\
      \midrule
      Literature
        & 82.5  & -  & 55.0   & -
        & 83.5  & -  & 58.1   & -
      \\
      \midrule
      \syssent
        & 50.0  & 37.0  & 70.0   & 17.5 %
        & 49.9  &  2.6  & 45.8   &  8.2 \\
      \midrule
      \syssentdoc 
        & 90.0  & 57.5  & 71.0  & 41.0     %
        & 90.3  & 13.9  & 75.3  & 32.4
      \\
      \sysdocboth
        & 90.0  & 60.5  & 73.5  & 43.0 %
        & 91.6  & 16.5  & 91.6  & 39.4
      \\
      \sysdocpar
        & 87.0  & 59.0  & 67.5  & 38.5       %
        & 88.3  & 16.6  & 88.0  & 39.6
      \\
      \sysdocbt 
        & 90.0  & 62.5  & 72.0  & 40.0      %
        & 95.0  & 16.9  & 87.3  & 39.6
      \\
      \bottomrule
    \end{tabular}
    \caption{Document constrastive test suites and their generative variants, where force-decoding the target context prefix is required (DiscEvalMT for \enfr, and part of GTWiC for \enru.
    In this setting, the document systems are all mostly indistinguishable.
    }
    \label{table:forced}
\end{table*}

\begin{table*}[pt]
    \centering
    \begin{tabular}{l|p{5.5in}}
    \toprule
docid &  t1\_h04bvl7 \\
src   &  Mal kurz gucken, mal kurz helfen, mal die Flasche bringen etc. \\
\syssent & Take a quick look, help for a moment, bring the bottle, etc. \\
\sysdocbt & Sometimes a short look, sometimes a short help, sometimes a bottle etc. \\
ref   &  Sometimes check-up on them, sometimes help, sometimes get the bottle, etc. \\
\midrule
docid  & braunschweiger-zeitung.de.35094 \\
src  &   "Wir waren einfach nicht da - weder im Angriff, noch in der Abwehr", räumte Rückraumspielerin Xenia Smits ein und forderte: "Wir müssen jetzt schnell nach vorne schauen". \\
\syssent &  "We just weren't there either in attack or defence," admitted centre-back Xenia Smits. \\
\sysdocbt & "We just weren't there - neither in attack nor in defence," admitted backroom player Xenia Smits, adding: "We have to look quickly forward now." \\
ref &     “We just weren’t in it - not in attack or defense,” said defender Xenia Smits, advising: “We just need to quickly turn towards the future”.\\
\midrule
docid  & en\_de\_CLIENT-01\_default\_2020-12-27-61\_second\_batch\_doc04 \\
src  &   Ihn so zu behalten ist für mich keine Option... \\
\syssent &  Keeping him like this isn't an option for me... \\
\sysdocbt & Keeping it like this is not an option for me... \\
ref  &   Keeping it the way it is isn’t an option for me. \\
    \bottomrule
    \end{tabular}
    \caption{Example translations from WMT22.
    Changes are often subtle, but demonstrate the contextual system's ability to capture repetitive style, handle compound sentences, and translate pronouns.}
    \label{table:examples}
\end{table*}

\subsection{Document-level metrics}

Next, we turn to the document-level contrastive and generative metrics described in \S~\ref{section:contrastive}--\ref{section:generative}.
We split the test suites into two groups: those where correct translation requires only source-side context (Table~\ref{table:source}), and those that evaluate target-side consistency (Table~\ref{table:forced}).

For generative document metrics, we took special care with \syssentdoc.
It was not trained with the separator token, so we do not use them when joining sentences for inference.
This means that we cannot reliably identify the payload (\S~\ref{section:method}), which complicates evaluation. %
We work around this with a heuristic for identifying the portion of the output to search for the pronoun.
We first (a) compute the percentage of tokens occupied by the source payload relative to the complete source input,
(b) apply that percentage to the target side, selecting tokens from the end of the output. %
Spot-checking suggests this to be a reasonable heuristic.

\paragraph{Source context}

Table~\ref{table:source} contains results for all four language pairs. %
Of these, German--English is a bit of an outlier, since it is focused on word-sense disambiguation.
This is a setting where we might expect document context to help, and indeed the dataset was designed to allow such experiments \citep{rios-gonzales-etal-2017-improving}.
But the high scores on the standard contrastive task suggest that the task is more-or-less solved; the fact that they are achieved by our sentence baseline suggests that document context was never necessary.
For the generative setting, results are also pretty comparable, with all systems performing more-or-less equivalently. %

For the other tasks and language pairs, we highlight an important pattern: with the contrastive metrics, the document systems improve as a block.
However, \emph{the generative metrics see their best results in the \sysdocbt system, often by a large margin}.
This is especially true for ContraPro and GenPro for \engdeu and \enfr.
Additionally, the \syssentdoc system \emph{improves} over the \syssent system when measured contrastively, but these gains are not reflected in the generative metric.
This raises a red flag, since we know this system has no generative document capacity.  %

\paragraph{Target context}

Table~\ref{table:forced} groups together document-level metrics that measure target-side consistency.
For these tests, the translation system is provided with the source document \emph{and} some target side context, and the system is tasked in its ability to resolve ambiguities that depend on both.
The task is therefore significantly more constrained and therefore likely easier.%

In this setting, the contrastive pattern is similar to that in Table~\ref{table:source}, with the numbers more-or-less unable to discriminate amongst document-level systems.
The generative metrics, too, are nearly identical across the systems.
We suspect that this may be resulting from providing the target-side context and force-decoding to it, which makes for a more constrained, easier task, perhaps closer to scoring than generation.
We investigate this more in Section~\ref{section:force-decode}.

\subsection{A closer look at GenPro}

\begin{table}[t]
    \centering
    \begin{tabular}{l|rr|rr}
    \toprule
      & \multicolumn{2}{c}{0}
      & \multicolumn{2}{c}{1+}
    \\
    & all & BT
    & all & BT
    \\
    \toprule
    all
    & 74.3 & 73.2
    & 47.7 & 70.5
    \\
    es
    & 86.2 & 81.2
    & 94.3 & 88.8
    \\
    sie
    & 72.9 & 73.6
    & 29.9 & 64.7
    \\
    er
    & 61.7 & 63.5
    & 20.7 & 58.8
    \\
    sie$\mid$er
    & 67.5 & 68.7
    & 25.2 & 61.8
    \\
    \bottomrule
    \end{tabular}
    \caption{Breakdown of ContraGen pronoun prediction accuracy by antecdent distance between two document systems: one (``all'') trained on docs from everywhere (\sysdocboth), and the other (``BT'') trained on docs only from BT data (\sysdocbt).
    The former has significantly-lower extra-sentential generative capacity.
    }
    \label{table:contragen}
\end{table}

In this section we look closer at the difference between the \sysdocboth and \sysdocbt \engdeu systems in Table~\ref{table:source}, which have similar ContraPro scores but divergent GenPro scores.
We wish to confirm in particular that the GenPro scores are capturing a real difference.

Table~\ref{table:contragen} provides a breakdown in performance between the two systems by antecedent distance and pronoun type.
The systems are essentially equivalent at distance=0, but quite divergent when the prediction requires looking into the sentence history.
Interestingly, we see that the gains are due to \sysdocbt's ability to correctly predict \emph{sie} and \emph{er}.
\sysdocboth is actually better at predicting \emph{es}.
This suggests that it simply overpredicts that pronoun, possibly due to its greater frequency in the training data.

\begin{table*}[t]
    \centering
    \begin{tabular}{l|p{5in}}
    \toprule

    \midrule
    src & You took the statue out? {\sep}No, it wasn't there.
    \\
    good & Sie haben die Statue herausgenommen? {\sep}Nein, {\color{green}\textbf{sie}} war nicht da. 
    \\
    bad & Du hast die Statue rausgeholt? {\sep}Nein, {\color{red}\textbf{er}} war nicht da.
    \\
    \midrule
    src & This creature here is covered by a coral.{\sep}It uses it as a protective cover and as a weapon.{\sep}It is hard like ceramic.
    \\
    good & Diese Kreatur hier ist von einer Koralle bedeckt. {\sep}Sie benutzt sie als Schutzhülle und als Waffe. {\sep}{\color{green}\textbf{Sie}} ist hart wie Keramik.
    \\
    bad &   Diese Kreatur hier ist von einer Koralle bedeckt. {\sep}Sie benutzt es als Schutzhülle und als Waffe. {\sep}{\color{red} \textbf{Es}} ist hart wie Keramik.
    \\
    \bottomrule
    \end{tabular}
    \caption{Examples of outputs rewarded/penalized by GenPro.}
    \label{table:contragen:examples}
\end{table*}

We note that GenPro may unfairly penalize a system that produces a correct sentence not containing the pronoun, or unfairly credit a system that happens to generate the pronoun in reference to different antecedent.
Small differences are therefore likely not significant, but spot-checking suggests to us that the large differences reported in Table~\ref{table:source} are meaningful.
\subsection{A closer look at target context}
\label{section:force-decode}

Why aren't the generative metrics in Table~\ref{table:forced} able to discriminate amongst document systems, whereas the generative metrics for the source-based translations in Table~\ref{table:source} were?
Many variables changed, so it's possible it has to do with differences in the languages (\enfr and \enru), or the datasets.
Another possible explanation is that force-decoding to target context is an unnatural task that pushes the model into inhospitable spaces \cite[\S~6.2]{post-vilar-2018-fast}, and that system performance here is more akin to contrastive evaluation, in which case it is  unlikely to be representative of its unconstrained generative behavior.

We tested this with a further change to the GenPro datasets (both \engdeu and \enfr).
Instead of providing only the source side, as in Table~\ref{table:source}, we also force-decode the target side up through the word preceding the pronoun.
We then compute the GenPro accuracy metric on the outputs.\footnote{We might also have looked at the distributions over the target vocabulary, but it was simpler to use the existing metric.}
The resulting situation is basically a discriminative setting.
The results can be found in Table~\ref{table:contrapro-forced}.

\begin{table}[ht]
    \centering
    \begin{tabular}{l|rr|rr}
    \toprule
      & \multicolumn{2}{c}{\engdeu} & \multicolumn{2}{c}{\enfr}
      \\
      system & free & forced & free & forced
      \\
    \midrule
      \sysdocboth 
      & 47.8  & 72.2  %
      & 62.5  & 74.7  %
      \\
      \sysdocpar  
      & 41.9  & 67.5
      & 60.4  & 72.0
      \\
      \sysdocbt   
      & 70.5  & 71.1
      & 77.3  & 75.7
      \\ 
     \bottomrule
    \end{tabular}
    \caption{GenPro scores, when force-decoding the target context up through the word prior to the pronoun.
    The \emph{free} column is repeated from Table~\ref{table:source} for ease of comparison.
    All document models are able to resolve the correct pronoun equally when provided forced context.
    }
    \label{table:contrapro-forced}
\end{table}

We see what we expected: the GenPro accuracies in the ``forced'' setting are much higher for \sysdocboth and \sysdocpar; they have basically been brought on par with \sysdocbt, whose results are basically unchanged.
The metric is no longer useful; only the fully generative setting exercises the model in a predictive way. 
\subsection{Model capacity}
\label{section:capacity}

\begin{table}[t]
    \centering
    \begin{tabular}{l|rrrr}
    \toprule
      arch/params
      & \small BLEU & \small COMET & \small C/Pro & \small G/Pro \\
    \midrule
  6/1k/146m	& 27.0	& 48.7	& 65.2 & 58.4 \\
  6/2k/171m	& 27.4	& 49.7	& 66.2 & 58.7 \\
  6/4k/221m	& 28.0  & 51.0	& 69.7 & 62.9 \\
  12/4k/297m	& 28.4	& 51.8	& 70.6 & 66.0 \\
  6/8k/322m	& 27.8	& 51.0	& 71.7 & 62.8 \\
  12/8k/448m	& 28.6	& 52.5	& 74.2 & 67.1 \\
  6/16k/523m  & 28.4  & 51.7  & 74.5 & 64.9 \\
  18/8k/574m	& 28.8	& 53.0	& 75.0 & 67.1 \\    
  12/16k/750m	& 28.9	& 52.8	& 75.8 & 68.5 \\
  18/16k/977m	& 29.3	& 53.3	& 75.5 & 69.4 \\
    \bottomrule
    \end{tabular}
    \caption{Model capacity (encoder layers / FFN size) for an EN-DE document model, ordered by param.\ count.
    Decoder depth is always 6 layers.
    Scores were computed on a checkpoint after ~30k updates.
    BLEU and COMET scores are on WMT21, translating as sentences.
    C/Pro is over the complete test set, while G/Pro is over only sentences with external anaphora.}
    \label{table:capacity}
\end{table}

Much work in investigating document-level machine translation has been limited to standard-size Transformer architectures (cf.\ \citet{zhang-etal-2018-improving,sun-etal-2022-rethinking,lopes-etal-2020-document}).
Yet it stands to reason that modeling longer-range phenomena will require increased model capacity, and in fact, the base model size we chose for our experiments (12 layer encoder, 16k FFN) reflects this.
Here, we provide more detail, varying two model parameters only: (i) the number of encoder layers, and (ii) the width of the model feed-forward layer (encoder and decoder side).
We keep all other parameters the same, including fixing the decoder depth to 6.
Focusing on changes to the encoder depth helps limit grid search and is justified by prior work showing that (relatively cheap) encoder layers can be traded for (relatively expensive) decoder layers with no penalty \cite{kasai-etal-2020-deep}.
We alternate between increasing the number of encoding layers, and increasing the dimension of the Transformer feed-forward layer.

Table~\ref{table:capacity} lists the results for English--German across our test suite. %
Unsurprisingly, all scores continue to rise, up to the wide 18-layer model.
Both increasing the number of encoder layers, and increasing the size of the FFN, contribute to better performance.
This suggests that the common approach of working with 6-layer Transformer base models is not enough for document-context MT.
There is more to gain by moving to larger models and likely, to larger datasets and sentence lengths, as well.
We leave further explorations to future work.

\subsection{Effect of context at inference}

\begin{figure}[t]
    \centering
    \includegraphics[width=0.48\textwidth]{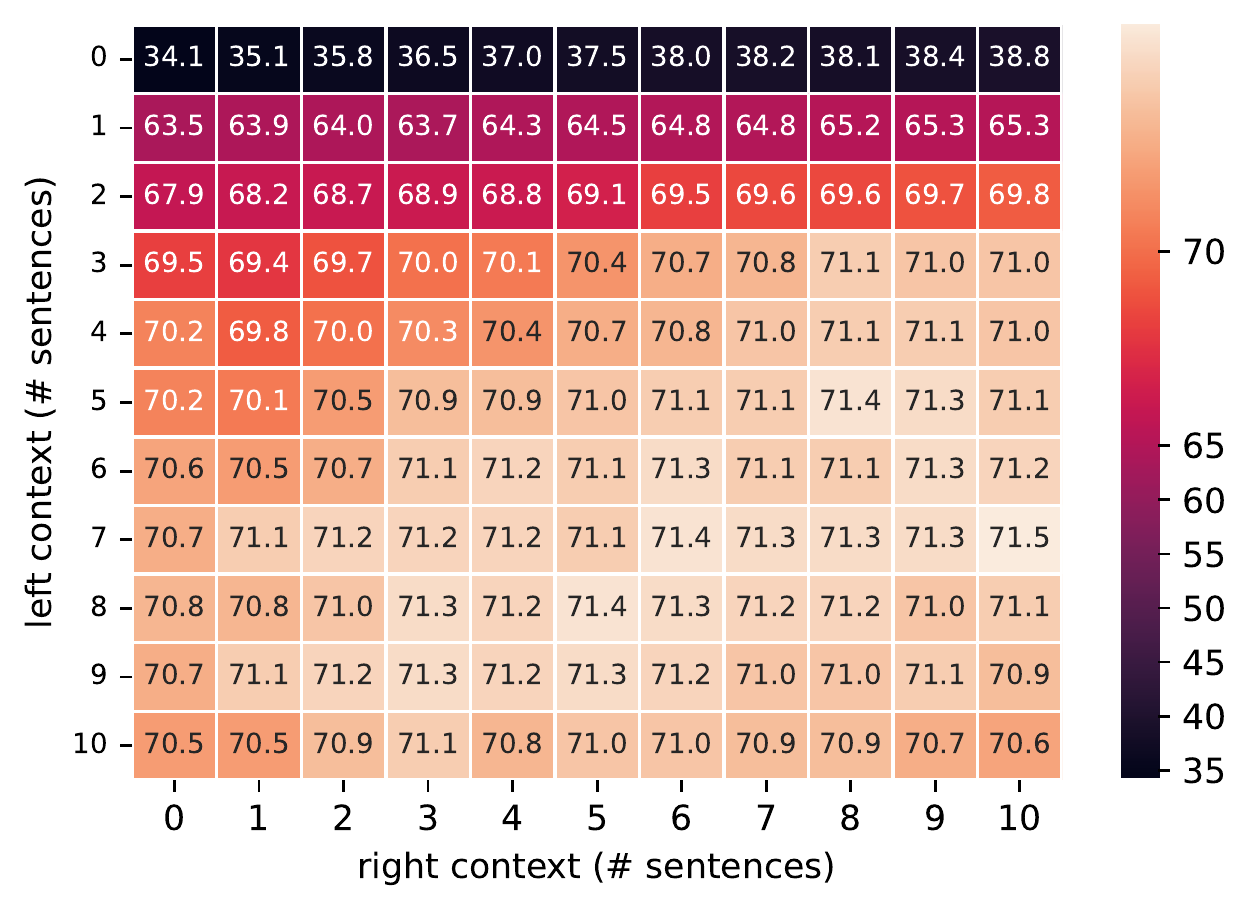}
    \caption{GenPro accuracies for EN-DE, reporting across all pronouns with extra-sentential anaphora.}
    \label{figure:context-sents}
\end{figure}

An important question is how much context is required to handle document-level phenomena.
The English--German ContraPro datasets nicely include extensive annotations that make it possible to answer this question for anaphora.
Figure~\ref{figure:context-sents} reports the effect of left and right sentence context on GenPro accuracies.
(Alternate views, including the effect when working within a constrained token budget, can be found in Appendix~\ref{appendix:genpro}).

A number of conclusions can be drawn.
Unsurprisingly, left context is critical; in fact, with zero left context, the task is essentially random.
Next, given a choice, it is nearly always better to add another sentence of left context than of right context.
However, the highest numbers are never reached without at least one sentence of right context.
Finally, gains continue to accrue as more context is added, up to a budget of seven or eight sentences.

Of course, it is unclear how these settings will generalize to other languages and for other document-level phenomena.
Future context is likely to be important as well for handling other phenomena \cite{wong-etal-2020-contextual}.

\subsection{Discourse-dense datasets}
\label{section:dense}

Table~\ref{table:bleu} (and Table~\ref{table:wmt21} in Appendix~\ref{appendix:wmt21}) show modest improvements when translating WMT test news test sets as documents, with document systems.
However, what is not clear is what the upper bound on performance is for document-level systems; in other words, how much unrealized gain is there that could have been addressed by contextual translation, specifically?  %
This is difficult to answer because we don't know how many document-level phenomena there are in these test sets, and in fact, we suspect there are relatively few.

As a means of assessing the question, we turn again to the ContraPro (\engdeu and \enfr) datasets, which we know to be extremely rich in one particular kind of document-level phenomena: pronoun selection.\footnote{OpenSubtitles is not in our training data.}
We take the positive examples along with their references as a normal test set (of size 12k).
We also create a second, shifted test set comprising the set of sentences \emph{after} each sentence in ContraPro, and call it ContraPro+1.
This second test set is likely to be significantly less rich in document phenomena than ContraPro, which gives us a baseline to compare system differences to.
We then compute COMET scores on these two test sets, translating their sentences both with our sentence system (\syssent, without context) and our document system (\sysdocbt, with context).\footnote{Documents have a maximum of 250 SPM tokens and 10 sentences.}

As we see in Table~\ref{table:contrapro-shifted}, the best document system (\sysdocbt) is \emph{much} better than the sentence baseline on ContraPro (+11.1 COMET for \engdeu, +4.4 for \enfr), suggesting that where document phenomena are rich, a document system is more helpful.
Next, for both language pairs, the gap also exists on the shifted test set (+3.3 \engdeu, +1.1 \enfr), but is much smaller.
Notably, the other document systems are close to \sysdocbt on the shifted test set (though still trailing), but somewhere in the middle on the dense test set.
Finally, the gaps are much tighter for \enfr than for \engdeu, which might suggest this dataset is less discourse-dense, or that the general task is easier for that language (two pronouns, instead of three).
Together, these facts suggest a challenge for the evaluation of document-level systems, which is the need to automatically identify sentences that require context to translate correctly, which is not an easy task \cite{bawden-etal-2018-detecting}.

\begin{table}[t]
    \centering
    \begin{tabular}{l|rr|rr}
    \toprule
      & \multicolumn{2}{c}{\engdeu} & \multicolumn{2}{c}{\enfr} \\
      system & C/Pro & \small shifted & C/Pro & \small shifted
      \\
    \midrule
      \syssent  & 21.4  & 31.4
      & 36.2  & 38.5
      \\
      \sysdocboth & 27.8 & 33.9
      & 38.4 & 39.2
      \\
      \sysdocpar & 26.0 & 34.1
      & 37.6 & 39.3
      \\
      \sysdocbt & 32.4  & 34.7
      & 40.6  & 39.6
      \\ 
      \midrule
      improvement & +11.1 & +3.3
      & +4.4 & +1.1
      \\
    \bottomrule
    \end{tabular}
    \caption{COMET (comet-wmt20-da) scores on the two datasets, the first (ContraPro) discourse-dense, the second (ContraPro+1) less so, for both \engdeu and \enfr.
    The gap between translating without and with context is much larger on the discourse-dense subset.}
    \label{table:contrapro-shifted}
\end{table}

\subsection{WMT experiment}
\label{section:wmt}

As noted above, a major limitation in the current sentence-level machine translation paradigm is the lack of document-level metadata on most parallel and even mono data. %
As a result, we were unable to perform the full set of system comparisons on open data. %
In a gesture towards repeatibility, we ground these experiments by training variants of our \syssent and \sysdocbt systems using a subset of WMT22 \engdeu data that has monolingual document annotations.

We use all available parallel data provided for WMT22 \cite{kocmi-etal-2022-findings}:\footnote{\url{https://statmt.org/wmt22/translation-task.html}} Europarl v10 \cite{koehn-2005-europarl}, Paracrawl v9 \cite{banon-etal-2020-paracrawl}, Common Crawl,\footnote{\url{https://commoncrawl.org/}} News Commentary, Wiki Titles v3, Tilde MODEL Corpus \cite{rozis-skadins-2017-tilde}, and Wikimatrix \cite{schwenk-etal-2021-wikimatrix}.
A few of these resources have document-level information, but we do not use any of it.
For monolingual data, the only data available with document metadata is News Crawl.\footnote{\url{https://data.statmt.org/news-crawl/de-doc/}}
We used all even years from 2008--2020, backtranslating it from German to English with an internal system.
Statistics for the data can be found in Table~\ref{table:wmtdata}.
No filtering is applied to any of it.
From this data, we train the only two of our systems supported by this setup: \syssent and \sysdocbt. 
These are trained for 40 virtual epochs each using the same settings described in Section~\ref{section:models}.

\begin{table}[t]
    \centering
    \begin{tabular}{l|rrr}
       \toprule
       source 
        & lines & docs & mean
        \\
       \midrule
       mono    
         & 311.2  &   14.1   &  21.9
       \\
       parallel
         & 297.6  &  0   & -
       \\
       \bottomrule
    \end{tabular}
    \caption{WMT training data statistics (lines and docs in millions). 
    The \emph{mean} column is the mean document length in sentences, excluding documents of length 1.}
    \label{table:wmtdata}
\end{table}

Results can be found in Table~\ref{table:wmtresults}. %
They are encouraging: we see the same pattern of improvement between \syssent and \sysdocbt, although the absolute numbers are lower.
Compared to our in-house data, the document metrics are even better for \syssent.

\begin{table}[t]
    \centering
    \begin{tabular}{l|rr|rr}
    \toprule
    & \multicolumn{2}{c|}{WMT22}
    \\
    system  & \small BLEU & \small COMET & \small C/Pro & \small G/Pro
    \\
    \midrule
    \syssent   & 35.8 & 60.6 & 50.8 & 35.3  \\
    \sysdocbt  & 35.8 & 59.4 & 83.4 & 64.3 \\
    \bottomrule
    \end{tabular}
    \caption{Metrics on the only two models we are able to build on public data (since there are insufficient document metadata for public parallel data).
    Similar patterns are observable to those seen in Table~\ref{table:bleu} and Table~\ref{table:source}.}
    \label{table:wmtresults}
\end{table}

\section{Related Work}
\label{section:related-work}

The dominance of the sentence-level paradigm in machine translation is not for lack of research effort.
A good early survey is \citet{maruf-etal-2019-survey}, who cover work with both RNN and Transformer frameworks along a rich taxonomy.
Here, we will attempt to survey recent state-of-the-art along the dimensions explored in this paper: architectures, data, and evaluation.
We build in many ways on prior work; part of our contribution is bringing a few key ideas together at once.

\paragraph{Architectures}
The Markov assumptions of statistical translation limited its ability to model long contexts.
There were attempts in the phrasal era \citep{gimenez-marquez-2007-context,carpuat-wu-2008-evaluation}, but the added complexity seemed to inhibit its adoption.
The transition to neural architectures was therefore a paradigm enabler for document translation.
Much work, including that with Transformers, has focused on separately encoding the context from the current sentence, in attempts to concentrate the relevant portions of the history and decrease sequence length.
This includes cache models \citep{tu-etal-2018-learning,kuang-etal-2018-modeling}, hierarchical attention \citep{miculicich-etal-2018-document}, separately encoding context \cite{voita-etal-2018-context,zhang-etal-2018-improving}, allowing attention across a batch of pseudo-documents \cite{wu-etal-2023-document}, encoding sentence position \cite{bao-etal-2021-g,lupo-etal-2023-encoding}, and sparse attention mechanisms \cite{guo-etal-2019-star}.
Another approach is post-processing approaches inspiring by automatic-post-editing but using document-level language models \cite{voita-etal-2019-context}.
\citet{yu-etal-2020-better} use Bayes' rule to factor translation and language modeling, translating sentences independently but using a document-level target language model to rerank candidates.

\citet{sun-etal-2022-rethinking} also proposed to stick with standard Transformer models.
They translate in the same ``doc-to-doc'' fashion that we use here, but use small datasets and models (6 layers) with no monolingual data, and generate training data with a ``multi-resolutional'' document-sampling approach that uses many different subsequences of each document.

Most work evaluates in small data settings, showing improvement on BLEU score on standard test sets or in contrastive settings.
We have not evaluated these models in our setting (large data, with generative metrics), which leaves open the question of whether they might still be of some help.
However, we suspect the added complexity is not necessary.
\paragraph{Data} 

There is very little parallel document-level data available.
Standard datasets with such information are OpenSubtitles \citep{lison-tiedemann-2016-opensubtitles2016}, IWSLT \citep{}, News Commentary, and Europarl \citep{koehn-2005-europarl}.
\citet{liu-zhang-2020-corpora} provide a nice survey, and release a small amount of government-crawled new data for Chinese--Portuguese.
The Conference on Machine Translation began releasing limited document-level data for DE-EN and CS-EN in 2019 \citep{barrault-etal-2019-findings}.

The limited amount of parallel document-level data has nonetheless produced a number of creative applications.
\citet{dobreva-etal-2020-document} incorporate finer-grained document structure using side constraints and the cache model of \citet{kuang-etal-2018-modeling}.
The idea to draw document data only from monolingual sources has also been tried.
\citet{voita-etal-2019-good} built a monolingual post-editing system that took the output of a baseline system and used it for document-level ``repair''. 
They found that it helped, but their models were small.
\citet{sugiyama-yoshinaga-2019-data} also used target-side data for backtranslation, evaluating in small-data settings with BLEU and contrastive metrics.
Our work differs by scaling this to very large datasets, and by showing that parallel data, as a whole, is actually harmful.

\paragraph{Metrics and Evaluation} 
A center point of document-level research is on metrics.
PROTEST \citep{guillou-hardmeier-2016-protest} was similar in spirit to our ContraGen.
They used hand-designed pronoun test cases and looked for the correct pronoun in the system output.
Failure cases were referred to humans for analysis.
\citet{laubli-etal-2018-machine} provided early evidence that document-level metrics would be helpful.
There has also been recent work in building automatic metrics that make use of context.
BlonDe \citep{jiang-etal-2022-blonde} was evaluated in Chinese--English and works by automatically identifying discourse-relevant phenomena in the output and comparing them to a reference, optionally combined with an n-gram fluency component.
Doc-COMET \citep{vernikos-etal-2022-embarrassingly} is simpler and builds sentence representations from context.
Both metrics are interesting but await deeper evaluation and we did not explore them in this paper.

\citet{vamvas-sennrich-2021-limits} have noted the problem with the disconnect between contrastive evaluation and generative ability for machine translation, but suggest using machine-generated minimal pairs that are closer to model distributions, and don't explore directly measuring generative ability.
\citet{bawden-etal-2018-detecting} note the difficulty of automatically identifying sentences that require context, which is relevant in light of our experiments in Section~\ref{section:dense} showing this to be important for model evaluation.

\section{Summary and Conclusions}

The field of machine translation can be described as being in a ``sentence-level rut''.
It is obvious that translation is not a sentence-level task, but the paradigm is so deeply ingrained into both research and production environments that it is hard to break out of, despite the clear and well-understood benefits.
The situation is also increasingly at odds with developments in the field, where large language models trained with standard architectures on long documents have completely overtaken nearly every leaderboard.

We identified three impediments to moving the field to contextual translation, and provided workable starting-point solutions to them all.
On the question of the correct architecture, we have shown the utility of standard Transformer models trained with sufficient capacity.
On the question of training data, we have shown that drawing document samples from backtranslated monolingual data alone suffices, and that parallel data, as a block, can actually be harmful.
And on the question of evaluation, we have shown that existing contrastive test beds do not discriminate document systems, but can be easily transformed into generative variants that do.
Furthermore, we have shown the importance of using ``discourse-dense'' datasets to evaluate document systems.

We see our approach here as a starting point, and many questions remain.
The fact that high-capacity Transformers work well does not preclude further improvements from more complex approaches.
Finally, perhaps the biggest unresolved issue is how to build scalable, trustworthy document metrics.
Code and data for many of our experiments and results can be downloaded from \repo.

\section*{Acknowledgments}

We would like to thank the authors of the contrastive test sets (particularly ContraPro) for their fantastic work creating well-designed, usable, and easily-extendable datasets, along with code.
Thanks also to Rachel Bawden, Salvador Mascarenhas, Christian Federmann, Tom Kocmi, Vikas Raunak, Arul Menezes, and Huda Khayrallah for helpful comments and feedback, and to Fai Sigolov for help with Russian.

\bibliography{anthology,custom}
\bibliographystyle{acl_natbib}

\appendix

\section{Results on WMT21}
\label{appendix:wmt21}

Table~\ref{table:wmt21} contains results on WMT21 \cite{akhbardeh-etal-2021-findings}.

\begin{table*}[t]
    \centering
    \begin{tabular}{ll|rr|rr|rr}
       \toprule
       \multicolumn{2}{c}{} 
       & \multicolumn{2}{c}{\textbf{DE$\rightarrow$EN}}
       & \multicolumn{2}{c}{\textbf{EN$\rightarrow$DE}}
       & \multicolumn{2}{c}{\textbf{EN$\rightarrow$RU}}
       \\
       \multicolumn{2}{c}{model}
       & BLEU & COMET
       & BLEU & COMET
       & BLEU & COMET
       \\
      \midrule
      \multicolumn{2}{c|}{Microsoft}
        &  34.3    & 62.1  %
        &  29.7    & 53.5  %
        &  29.2    & 58.9  %
      \\
      \midrule
     \multirow{4}{*}{\rotatebox{90}{sent-level}}
      & \syssent
        & 33.0 & 60.5 
        & 28.5 & 52.6
        & 29.7 & 63.0
      \\
      & \sysdocboth 
        & 33.1 & 60.6
        & 29.2 & 53.2
        & 29.7 & 63.5
        \\
      & \sysdocpar
        & 32.2 & 60.2
        & 28.5 & 53.1
        & 30.0 & 63.2
      \\
      & \sysdocbt
        & 33.2 & 60.7
        & 29.0 & 53.2  %
        & 30.0 & 63.3
        \\
      \midrule
      \multirow{3}{*}{\rotatebox{90}{doc-level}}
      & \sysdocboth 
        & 34.7  & 61.4
        & 29.2  & 51.3
        & 29.6  & 61.3
        \\
      & \sysdocpar
        & 34.3  & 60.6
        & 29.2  & 51.8
        & 29.1  & 60.4
      \\
      & \sysdocbt
        & 31.3  & 60.5
        & 28.8  & 53.3
        & 30.0  & 63.2
      \\      
      \bottomrule
    \end{tabular}
    \caption{Metric scores on WMT21 test sets when translating with and without document context.
    Numbers within a column are comparable.
    \syssentdoc misuses the sentence-based model by feeding it documents in a zero-shot fashion.
    }
    \label{table:wmt21}
\end{table*}

\section{The effect of context on GenPro with a token budget}
\label{appendix:genpro}

In production settings, we do not have the ability to include arbitrary amounts of sentence context due to the latency implications.
Instead, a constraint is placed on the maximum number of tokens in any single input.
We therefore might wonder how the results in Figure~\ref{figure:context-sents} compare when operating under this setting.

\begin{figure}[t]
    \centering
    \includegraphics[width=0.5\textwidth]{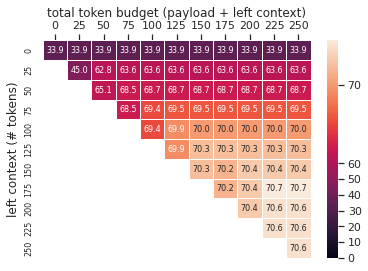}
    \caption{Token context. 
    GenPro accuracies for EN-DE as a function of the number of the total token budget (columns), including the payload, and the number of those tokens allocated to the left context (rows).
    Only complete sentences are adding to context. 
    Leftover tokens are allocated to the right context.
    }
    \label{figure:context-tokens}
\end{figure}

We experiment with a token budget setting of the form $(n, l)$, where $n$ is the maximum number of tokens, and $l$ is the number of tokens available for the left context.
To construct each input, we begin with the payload $p$ (\S~\ref{section:method}).
We then construct the left-sentence context $c_l$ using the remaining $n-|p|$ tokens, always including only completely sentences.
The remaining $n-|c_l|-|p|$ tokens are then available for right-sentence context.
Sentences are joined with the \sep token and translated as a single unit, then evaluated with GenPro.
The heatmap in Figure~\ref{figure:context-tokens} shows similar results to those in Figure~\ref{figure:context-sents}.

\end{document}